\pdfoutput=1

\documentclass[11pt]{article}

\usepackage{emnlp2021}

\usepackage{times}
\usepackage{latexsym}

\usepackage[T1]{fontenc}

\usepackage[utf8]{inputenc}
\usepackage{enumitem}
\usepackage{microtype}

\usepackage{xcolor, graphicx}
\usepackage[export]{adjustbox}
\usepackage{subcaption}
\usepackage{multirow}
\usepackage{amsfonts}
\usepackage{amsmath}
\usepackage{enumitem}

\title{PASTE: A Tagging-Free Decoding Framework Using Pointer Networks for Aspect Sentiment Triplet Extraction}

\author{Rajdeep Mukherjee\thanks{Equal contribution}, Tapas Nayak\footnotemark[1], Yash Butala, \\\textbf{Sourangshu Bhattacharya \and Pawan Goyal} \\
        Department of Computer Science and Engineering, IIT Kharagpur, India \\ \texttt{\{rajdeep1989, yashbutala\}@iitkgp.ac.in, tnk02.05@gmail.com,} \\ \texttt{\{sourangshu, pawang\}@cse.iitkgp.ac.in}}
        
\begin{document}
\maketitle
\begin{abstract}
Aspect Sentiment Triplet Extraction (ASTE) deals with extracting \emph{opinion triplets}, consisting of an opinion target or aspect, its associated sentiment, and the corresponding opinion term/span explaining the rationale behind the sentiment. Existing research efforts are majorly \emph{tagging}-based. Among the methods taking a \emph{sequence tagging} approach, some fail to capture the strong interdependence between the three \emph{opinion factors}, whereas others fall short of identifying triplets with overlapping aspect/opinion spans. A recent \emph{grid tagging} approach on the other hand fails to capture the span-level semantics while predicting the sentiment between an aspect-opinion pair. Different from these, we present a \emph{tagging-free} solution for the task, while addressing the limitations of the existing works. We adapt an encoder-decoder architecture with a \emph{Pointer Network}-based decoding framework that generates an entire opinion triplet at each time step thereby making our solution end-to-end. Interactions between the aspects and opinions are effectively captured by the decoder by considering their entire detected spans while predicting their connecting sentiment. Extensive experiments on several benchmark datasets establish the better efficacy of our proposed approach, especially in \emph{recall}, and in predicting multiple and aspect/opinion-overlapped triplets from the same review sentence. We report our results both with and without BERT and also demonstrate the utility of domain-specific BERT \emph{post-training} for the task.
\end{abstract}

\section{Introduction}\label{sec:intro}
Aspect-based Sentiment Analysis (ABSA) is a broad umbrella of several fine-grained sentiment analysis tasks, and has been extensively studied since its humble beginning in \emph{SemEval 2014} \citep{pontiki-etal-2014-semeval}. Overall, the task revolves around automatically extracting the opinion targets or aspects being discussed in review sentences, along with the sentiments expressed towards them. Early efforts on \emph{Aspect-level Sentiment Classification} \citep{AAAI18-ASC, li-etal-2018-transformation, xue-li-2018-aspect} focus on predicting the sentiment polarities for given aspects. However, in a real-world scenario, aspects may not be known a-priori. Works on \emph{End-to-End ABSA} \citep{li2019unified, he-etal-2019-interactive, chen-qian-2020-relation} thus focus on extracting the aspects as well as the corresponding sentiments simultaneously. These methods do not however capture the reasons behind the expressed sentiments, which could otherwise provide valuable clues for more effective extraction of aspect-sentiment pairs. 

\begin{table}[t]
    \centering
    \resizebox{\linewidth}{!}{
    \begin{tabular}{|l|l|}
        \hline
        Sent 1: & \begin{tabular}[c]{@{}l@{}}The \textcolor{blue}{film} was \textcolor{teal}{good} , but \textcolor{red}{could have been better} . \end{tabular} \\ 
        \hline
        Triplets & \begin{tabular}[c]{@{}l@{}}{[}Aspect ; Opinion ; Sentiment{]} \\ 
        (1) \textcolor{blue}{film} ; good ; \textcolor{teal}{positive} \\ 
        (2) \textcolor{blue}{film} ; could have been better ; \textcolor{red}{negative} \end{tabular} \\ 
        \hline
        Sent 2: & \begin{tabular}[c]{@{}l@{}}The \textcolor{blue}{weather} was \textcolor{red}{gloomy} , but the \textcolor{blue}{food} was \textcolor{teal}{tasty} . \end{tabular} \\ 
        \hline
        Triplets & \begin{tabular}[c]{@{}l@{}}
        (1) \textcolor{blue}{weather} ; gloomy ; \textcolor{red}{negative} \\ 
        (2) \textcolor{blue}{food} ; tasty ; \textcolor{teal}{positive} \end{tabular} \\ 
        \hline
    \end{tabular}
    }
    \caption{Examples of Aspect-Opinion-Sentiment triplets (\textit{opinion triplets}) present in review sentences.}
    \label{tab:example}
\end{table}

Consider the two examples shown in Table \ref{tab:example}. For the first sentence, the sentiment associated with the aspect \emph{\textcolor{blue}{film}}, changes depending on the connecting opinion phrases; \emph{\textcolor{teal}{good}} suggesting a \textcolor{teal}{positive} sentiment, and \emph{\textcolor{red}{could have been better}} indicating a \textcolor{red}{negative} sentiment. Hence, simply extracting the pairs \emph{film-positive}, and \emph{film-negative} without additionally capturing the reasoning phrases may confuse the reader. For the second sentence, the opinion term \emph{\textcolor{red}{gloomy}} has a higher probability of being associated with \emph{\textcolor{blue}{weather}}, than with \emph{\textcolor{blue}{food}}. We therefore observe that the three elements or \emph{opinion factors} of an \emph{opinion triplet} are strongly inter-dependent. In order to offer a complete picture of \emph{what} is being discussed, \emph{how} is the sentiment, and \emph{why} is it so, \citep{Peng2020KnowingWH} defined the task of \textbf{A}spect \textbf{S}entiment \textbf{T}riplet \textbf{E}xtraction (\textbf{ASTE}). Given an opinionated sentence, it deals with extracting all three elements: the aspect term/span, the opinion term/span, and the connecting sentiment in the form of \emph{opinion triplets} as shown in Table \ref{tab:example}. It is to be noted here that a given sentence might contain multiple triplets, which may further share aspect or opinion spans (For e.g., the two triplets for Sent. 1 in Table \ref{tab:example} share the aspect \emph{\textcolor{blue}{film}}). An efficient solution for the task must therefore be able to handle such challenging data points.


\newcite{Peng2020KnowingWH} propose a two-stage pipeline framework. In the first stage, they extract aspect-sentiment pairs and opinion spans using two separate sequence-tagging tasks, the former leveraging a \emph{unified tagging} scheme proposed by \citep{li2019unified}, and the later based on \emph{BIEOS}\footnote{BIOES is a commonly used tagging scheme for sequence labeling tasks, and denotes “begin, inside, outside, end and single” respectively.} tagging scheme. In the second stage, they pair up the extracted aspect and opinion spans, and use an MLP-based classifier to determine the validity of each generated triplet. \newcite{otemtl-zhang-etal-2020-multi-task} propose a multi-task framework to jointly detect aspects, opinions, and sentiment dependencies. Although they decouple the sentiment prediction task from aspect extraction, they use two separate sequence taggers (\emph{BIEOS}-based) to detect the aspect and opinion spans in isolation before predicting the connecting sentiment. Both these methods however break the interaction between aspects and opinions during the extraction process. While the former additionally suffers from error propagation problem, the latter, relying on word-level sentiment dependencies, cannot guarantee sentiment consistency over multi-word aspect/opinion spans. 

\newcite{jet-xu-etal-2020-position} propose a novel position-aware tagging scheme (extending \emph{BIEOS} tags) to better capture the interactions among the three \emph{opinion factors}. One of their model variants however cannot detect aspect-overlapped triplets, while the other cannot identify opinion-overlapped triplets. Hence, they need an ensemble of two variants to be trained for handling all cases. \newcite{wu-etal-2020-grid} try to address this limitation by proposing a novel grid tagging scheme-based approach. However, they end up predicting the relationship between every possible word pair, irrespective of how they are syntactically connected, thereby impacting the span-level sentiment consistency guarantees.


Different from all these tagging-based methods, we propose to investigate the utility of a \textbf{tagging-free} scheme for the task. Our innovation lies in formulating ASTE as a structured prediction problem. Taking motivation from similar sequence-to-sequence approaches proposed for \emph{joint entity-relation extraction} \citep{Nayak2020EffectiveMO, chen-etal-2021-jointly}, \emph{semantic role labeling} \citep{Fei-SRL-2021} etc., we propose \textbf{PASTE}, a Pointer Network-based encoder-decoder architecture for the task of ASTE. The pointer network effectively captures the aspect-opinion interdependence while detecting their respective spans. The decoder then learns to model the span-level interactions while predicting the connecting sentiment. An entire \textit{opinion triplet} is thus decoded at every time step, thereby making our solution end-to-end. We note here however, that the aspect and opinion spans can be of varying lengths, which makes the triplet decoding process challenging. For ensuring uniformity, we also propose a position-based representation scheme to be suitably exploited by our proposed architecture. Here, each \emph{opinion triplet} is represented as a 5-point tuple, consisting of the start and end positions of the aspect and opinion spans, and the sentiment (POS/NEG/NEU) expressed towards the aspect. To summarize our contributions:
\begin{itemize}[leftmargin=*]
    \item We present an end-to-end \emph{tagging-free} solution for the task of ASTE that addresses the limitations of previous tagging-based methods. Our proposed architecture, \textbf{PASTE}, not only exploits the aspect-opinion interdependence during the span detection process, but also models the span-level interactions for sentiment prediction, thereby truly capturing the inter-relatedness between all three elements of an \emph{opinion triplet}.
    \item We propose a position-based scheme to uniformly represent an opinion triplet, irrespective of varying lengths of aspect and opinion spans.
    \item Extensive experiments on the ASTE-Data-V2 dataset \cite{jet-xu-etal-2020-position} establish the overall superiority of \emph{PASTE} over strong state-of-the-art baselines, especially in predicting multiple and/or overlapping triplets. We also achieve significant (15.6\%) recall gains in the process.
\end{itemize}

\section{Our Approach}\label{sec:approach}
\begin{figure*}
    \centering
    \resizebox{\linewidth}{!}{
    \includegraphics{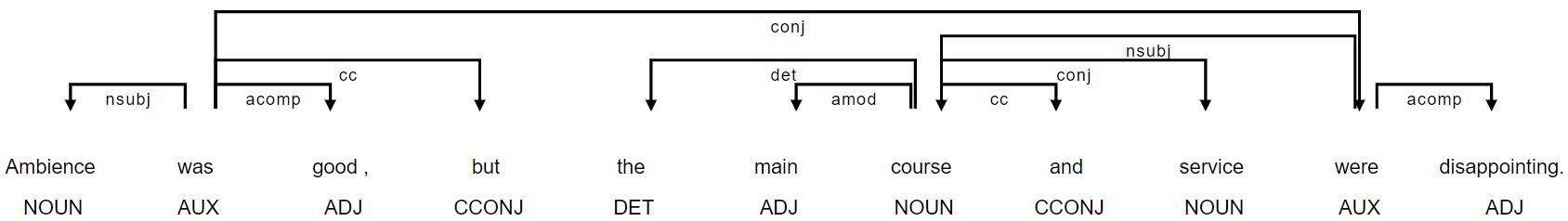}
    }
    \caption{Dependency Parse Tree for the example review sentence in Table \ref{tab:triplet_rep}}
    \label{fig:dep_tree}
\end{figure*}

\begin{table*}
    \centering
    \resizebox{0.9\linewidth}{!}{
    \begin{tabular}{p{0.22\linewidth} | p{0.78\linewidth}}
        \hline
        \bf{Sentence} & Ambience was good , but the main course and service were disappointing . \\
        \hline
        \bf{Target Triplets} & (0 0 2 2 POS) \quad (6 7 11 11 NEG) \quad (9 9 11 11 NEG) \\
        \hline
        \bf{Overlapping Triplets} & (6 7 11 11 NEG) \quad (9 9 11 11 NEG) \\
        \hline
    \end{tabular}
    }
    \caption{Triplet representation for Pointer-network based decoding}
    \label{tab:triplet_rep}
\end{table*}
Given the task of ASTE, our objective is to jointly extract the three elements of an opinion triplet, i.e., the aspect span, its associated sentiment, and the corresponding opinion span, while modeling their interdependence. Towards this goal, we first introduce our triplet representation scheme, followed by our problem formulation. We then present our Pointer Network-based decoding framework, \textbf{PASTE}, and finally discuss a few model variants. Through exhaustive experiments, we investigate the utility of our approach and present a performance comparison with strong state-of-the-art baselines.

\subsection{Triplet Representation}\label{subsec:triplet_rep}
In order to address the limitations of \emph{BIEOS} tagging-based approaches and to facilitate joint extraction of all three elements of an \emph{opinion triplet}, we represent each triplet as a 5-point tuple, consisting of the start and end positions of the aspect span, the start and end positions of the opinion span, and the sentiment (POS/NEG/NEU) expressed towards the aspect. This allows us to model the relative context between an aspect-opinion pair which is not possible if they were extracted in isolation. It further helps to jointly extract the sentiment associated with such a pair. An example sentence with triplets represented under the proposed scheme is shown in Table \ref{tab:triplet_rep}. As may be noted, such a scheme can easily represent triplets with overlapping aspect or opinion spans, possibly with varying lengths.

\subsection{Problem Formulation}\label{subsec:problem}
To formally define the ASTE task, given a review sentence $s = \{w_1, w_2, ..., w_n\}$ with $n$ words, our goal is to extract a set of opinion triplets $T = \{t_i \: | \: t_i = [(s^{ap}_i, e^{ap}_i), (s^{op}_i, e^{op}_i), senti_i]\}^{|T|}_{i=1}$, where $t_i$ represents the $i^{th}$ triplet and $|T|$ represents the length of the triplet set. For the $i^{th}$ triplet, $s^{ap}_i$ and $e^{ap}_i$ respectively denote the start and end positions of its constituent aspect span, $s^{op}_i$ and $e^{op}_i$ respectively denote the start and end positions of its constituent opinion span, and $senti_i$ represents the sentiment polarity associated between them. Here, $senti_i \in \{POS, NEU, NEU\}$, where $POS$, $NEG$, and $NEU$ respectively represent the \emph{positive}, \emph{negative}, and \emph{neutral} sentiments.

\begin{figure*}[ht]
    \centering
    \resizebox{0.74\textwidth}{!}{
    \includegraphics{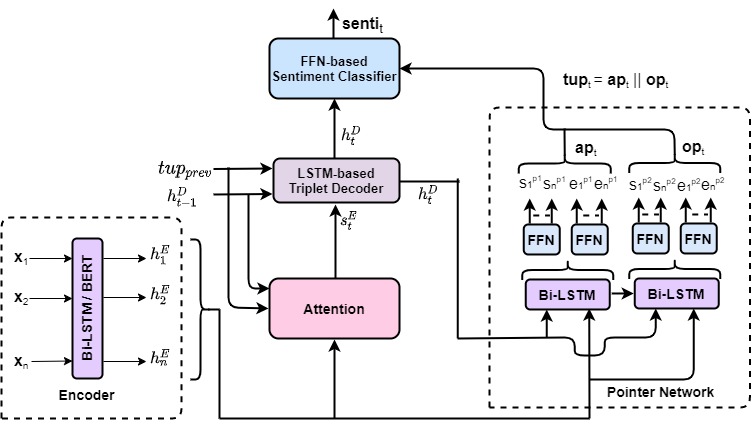}
    }
    \caption{Model architecture of PASTE, a Pointer Network-based decoding framework for ASTE.}
    \label{fig:model}
\end{figure*}

\subsection{The PASTE Framework}\label{subsec:framework}
We now present \textbf{PASTE}, our \textbf{P}ointer network-based decoding framework for the task of \textbf{A}spect \textbf{S}entiment \textbf{T}riplet \textbf{E}xtraction. Figure \ref{fig:model} gives an overview of our proposed architecture.


\subsubsection{Sentence Encoder}\label{subsubsec:encoder}
As previously motivated, the association between an aspect, an opinion, and their connecting sentiment is highly contextual. This factor is more noteworthy in sentences containing multiple triplets with/without varying sentiment polarities and/or overlapping aspect/opinion spans. Long Short Memory Networks (or LSTMs) \citep{lstm} are known for their context modeling capabilities. Similar to \citep{Nayak2020EffectiveMO, chen-etal-2021-jointly}, we employ a Bi-directional LSTM (Bi-LSTM) to encode our input sentences. We use pre-trained word vectors of dimension $d_w$ to obtain the word-level features. We then note from Figure \ref{fig:dep_tree} that aspect spans are often characterized by \textit{noun phrases}, whereas opinion spans are often composed of \textit{adjective phrases}. Referring to the dependency tree in the same figure, the aspect and the opinion spans belonging to the same opinion triplet are often connected by the same \textit{head word}. These observations motivate us to use both part-of-speech (POS) and dependency-based (DEP) features for each word. 

More specifically, we use two embedding layers, $E_{pos} \in \mathbb{R}^{\mathrm{|POS|} \: \times \: d_{pos}}$, and $E_{dep} \in \mathbb{R}^{\mathrm{|DEP|} \: \times \: d_{dep}}$ to obtain the POS and DEP-features of dimensions $d_{pos}$ and $d_{dep}$ respectively, with $\mathrm{|POS|}$ and $\mathrm{|DEP|}$ representing the length of POS-tag and DEP-tag sets over all input sentences. All three features are concatenated to obtain the input vector representation $\boldsymbol{\mathrm{x_i}} \in \mathbb{R}^{d_w + d_{pos} + d_{dep}}$ corresponding to the $i^{th}$ word in the given sentence $S = \{w_1, w_2, ..., w_n\}$. The vectors are passed through the Bi-LSTM to obain the contextualized representations $\boldsymbol{\mathrm{h_i^E}} \in \mathbb{R}^{d_h}$. Here, $d_h$ represents the hidden state dimension of the triplet generating LSTM decoder as detailed in the next section. Accordingly, the hidden state dimension of both the forward and backward LSTM of the Bi-LSTM encoder are set to $d_h / 2$.

For the BERT-based variant of our model, Bi-LSTM gets replaced by BERT \citep{devlin-etal-2019-bert} as the sentence encoder. The pre-trained word vectors are accordingly replaced by BERT token embeddings. We now append the POS and DEP features vectors to the 768-dim. token-level outputs from the final layer of BERT. 

\subsubsection{Pointer Network-based Decoder}\label{subsubsec:decoder}
Referring to Figure \ref{fig:model}, opinion triplets are decoded using an LSTM-based \textit{Triplet Decoder}, that takes into account the history of previously generated pairs/tuples of aspect and opinion spans, in order to avoid repetition. At each time step $t$, it generates a hidden representation $\boldsymbol{\mathrm{h_t^D}} \in \mathbb{R}^{d_h}$ that is used by the two Bi-LSTM + FFN-based \textit{Pointer Networks} to respectively predict the aspect and opinion spans, while exploiting their interdependence. The tuple representation $tup_t$ thus obtained is concatenated with $\boldsymbol{\mathrm{h_t^D}}$ and passed through an FFN-based \textit{Sentiment Classifier} to predict the connecting sentiment, thereby decoding an entire opinion triplet at the $t^{th}$ time step. We now elaborate each component of our proposed decoder framework in greater depth:

\subsubsection*{Span Detection with Pointer Networks}
Our pointer network consists of a Bi-LSTM, with hidden dimension $d_p$, followed by two feed-forward layers (FFN) on top to respectively predict the start and end locations of an entity span. We use two such pointer networks to  produce a tuple of hidden vectors corresponding to the aspect and opinion spans of the triplet to be decoded at time step $t$. We concatenate $\boldsymbol{\mathrm{h_t^D}}$ with each of the encoder hidden state vectors $\boldsymbol{\mathrm{h_i^E}}$ and pass them as input to the first Bi-LSTM. The output hidden state vector corresponding to the $i^{th}$ token of the sentence thus obtained is simultaneously fed to the two FFNs with \textit{sigmoid} to generate a pair of scores in the range of 0 to 1. After repeating the process for all tokens, the normalized probabilities of the $i^{th}$ token to be the start and end positions of an \textit{aspect span} ($s_i^{p_1}$ and $e_i^{p_1}$ respectively) are obtained using \textit{softmax} operations over the two sets of scores thus generated. Here $p_1$ refers to the first pointer network. Similar scores corresponding to the \textit{opinion span} are obtained using the second pointer network, $p_2$; difference being that apart from $\boldsymbol{\mathrm{h_t^D}}$, we also concatenate the output vectors from the first Bi-LSTM with encoder hidden states $\boldsymbol{\mathrm{h_i^E}}$ and pass them as input to the second Bi-LSTM. This helps us to model the interdependence between an aspect-opinion pair. These scores are used to obtain the hidden state representations $ap_t \in \mathbb{R}^{2d_p}$ and $op_t \in \mathbb{R}^{2d_p}$ corresponding to the pair of aspect and opinion spans thus predicted at time step $t$. We request our readers to kindly refer to the appendix for more elaborate implementation details.

Here we introduce the term \textit{generation direction} which refers to the order in which we generate the hidden representations for the two entities, i.e. aspect and opinion spans. This allows us to define two \textbf{variants} of our model. The variant discussed so far uses $p_1$ to detect the aspect span before predicting the opinion span using $p_2$, and is henceforth referred to as \textbf{PASTE-AF} (AF stands for \textit{aspect first}). Similarly, we obtain the second variant \textbf{PASTE-OF} (\textit{opinion first}) by reversing the \textit{generation direction}. The other two components of our model remain the same for both the variants.

\begin{table*}
    \centering
    \resizebox{\linewidth}{!}{
    \begin{tabular}{|l|c|c|c|c|c|c|c|c|c|c|c|c|c|c|c|}
        \hline
        \multirow{2}{*}{\textbf{Dataset}} & \multicolumn{3}{c|}{\textbf{14Lap}} &
        \multicolumn{3}{c|}{\textbf{14Rest}} & \multicolumn{3}{c|}{\textbf{15Rest}} & \multicolumn{3}{c|}{\textbf{16Rest}} & 
        \multicolumn{3}{c|}{\textbf{Restaurant (All)}} \\
        \cline{2-16}
        & \textbf{\# Pos.} & \textbf{\# Neg.} & \textbf{\# Neu.} & \textbf{\# Pos.} & \textbf{\# Neg.} & \textbf{\# Neu.} & \textbf{\# Pos.} & \textbf{\# Neg.} & \textbf{\# Neu.} & \textbf{\# Pos.} & \textbf{\# Neg.} & \textbf{\# Neu.} & \textbf{\# Pos.} & \textbf{\# Neg.} & \textbf{\# Neu.} \\
        \hline
        \textbf{Train} & 817 & 517 & 126 & 1692 & 480 & 166 & 783 & 205 & 25 & 1015 & 329 & 50 & 3490 & 1014 & 241 \\
        \textbf{Dev} & 169 & 141 & 36 & 404 & 119 & 54 & 185 & 53 & 11 & 252 & 76 & 11 & 841 & 248 & 76 \\
        \textbf{Test} & 364 & 116 & 63 & 773 & 155 & 66 & 317 & 143 & 25 & 407 & 78 & 29 & 1497 & 376 & 120 \\
        \hline
    \end{tabular}
    }
    \caption{ASTE-Data-V2 Statistics: \# Triplets with various sentiment polarities}
    \label{tab:ASTE-V2-trip}
\end{table*}

\begin{table*}
    \centering
    \resizebox{\linewidth}{!}{
    \begin{tabular}{|l|c|c|c|c|c|c|c|c|c|c|}
        \hline
        \multirow{2}{*}{\textbf{Dataset}} &
        \multicolumn{5}{c|}{\textbf{Laptop}} &
        \multicolumn{5}{c|}{\textbf{Restaurant}} \\
        \cline{2-11}
        & \textbf{Single} & \textbf{Multi} & \textbf{MultiPol} & \textbf{Overlap} & \textbf{\# Sent.} & \textbf{Single} & \textbf{Multi} & \textbf{MultiPol} & \textbf{Overlap} & \textbf{\# Sent.}\\
        \hline
        \textbf{Train} & 545 & 361 & 47 & 257 & 906 & 1447 & 1281 & 205 & 731 & 2728 \\
        \textbf{Dev} & 133 & 86 & 10 & 59 & 219 & 347 & 321 & 45 & 197 & 668 \\
        \textbf{Test} & 184 & 144 & 18 & 97 & 328 & 608 & 532 & 71 & 317 & 1140 \\
        \hline
        \textbf{Total} & \bf{862} & \bf{591} & \bf{75} & \bf{413} & \bf{1453} & \bf{2402} & \bf{2134} & \bf{321} & \bf{1245} & \bf{4536} \\
        \hline
        
    \end{tabular}
    }
    \caption{Statistics of \emph{Laptop} and \emph{Restaurant} datasets from ASTE-Data-V2: \textit{Single} and \textit{Multi} respectively represent \# sentences with single and multiple triplets. \textit{MultiPol} and \textit{Overlap} are subsets of \textit{Multi}. \textit{MultiPol} representing \# sentences containing at least two triplets with different sentiment polarities. \textit{Overlap} represents \# sentences with aspect/opinion overlapped triplets. \# Sent. represents the total no. of sentences overall.}
    \label{tab:ASTE-V2-split}
\end{table*}

\subsubsection*{Triplet Decoder and Attention Modeling}
The decoder consists of an LSTM with hidden dimension $d_h$ whose goal is to generate the sequence of opinion triplets, $T$, as defined in Section \ref{subsec:problem}. Let $tup_t = \mathrm{ap_t} \mathbin\Vert \mathrm{op_t} \:;\: \mathrm{tup_t} \in \mathbb{R}^{4d_p}$ denote the tuple (aspect, opinion) representation obtained from the pointer networks at time step $t$. Then, $tup_{prev} = \sum_{j < t} tup_j \:;\: tup_0 = \Vec{0} \in \mathbb{R}^{4d_p}$ represents the cumulative information about all tuples predicted before the current time step. We obtain an attention-weighted context representation of the input sentence at time step $t$ ($s_t^E \in \mathbb{R}^{d_h}$) using \newcite{BadhanauAttention} \textit{Attention}\footnote{Please refer to the appendix for implementation details.}. In order to prevent the decoder from generating the same tuple again, we pass $tup_{prev}$ as input to the LSTM along with $s_t^E$ to generate $\boldsymbol{\mathrm{h_t^D}} \in \mathbb{R}^{d_h}$, the hidden representation  for predicting the triplet at time step $t$:
\begin{equation}
    \boldsymbol{\mathrm{h_t^D}} = \boldsymbol{\mathrm{LSTM}}(s_t^E \mathbin\Vert tup_{prev} \:,\: \boldsymbol{\mathrm{h_{t-1}^D}}) \nonumber
\end{equation}

\subsubsection*{Sentiment Classifier}
Finally, we concatenate $tup_t$, with $\boldsymbol{\mathrm{h_t^D}}$ and pass it through a feed-forward network with softmax to generate the normalized probabilities over $\{POS, NEG, NEU\} \cup \{NONE\}$, thereby predicting the sentiment label $senti_t$ for the current triplet. Interaction between the entire predicted spans of aspect and opinion is thus captured for sentiment identification. Here $POS, NEG, NEU$ respectively represent the positive, negative, and neutral sentiments. $NONE$ is a dummy sentiment that acts as an implicit stopping criteria for the decoder. During training, once a triplet with sentiment $NONE$ is predicted, we ignore all subsequent predictions, and none of them contribute to the loss. Similarly, during inference, we ignore any triplet predicted with the $NONE$ sentiment.

\subsubsection{Training}\label{subsubsec:training}
For training our model, we minimize the sum of negative log-likelihood loss for classifying the sentiment and the four pointer locations corresponding to the aspect and opinion spans:
\begin{multline*}
    \mathcal{L} = - \frac{1}{M \times J} \sum_{m=1}^{M} \sum_{j=1}^{J} [log (s_{ap,j}^m \cdot e_{ap,j}^m) \\ + log (s_{op,j}^m \cdot e_{op,j}^m) + log (sen_j^m)]
\end{multline*}
\noindent Here, $m$ represents the $m^{th}$ training instance with $M$ being the batch size, $j$ represents the $j^{th}$ decoding time step with $J$ being the length of the longest target sequence among all instances in the current batch. $s_p, e_p; \;p \in \{ap, op\}$ and $sen$ respectively represent the softmax scores corresponding to the true start and end positions of the aspect and opinion spans and their associated true sentiment label.

\subsubsection{Inferring The Triplets}\label{subsubsec:inference}
Let $s_i^{ap}, e_i^{ap}, s_i^{op}, e_i^{op};\: i \in [1,n] $ represent the obtained pointer probabilities for the $i^{th}$ token in the given sentence (of length $n$) to be the start and end positions of an aspect span and opinion span respectively. First, we choose the start ($j$) and end ($k$) positions of the aspect span with the constraint $1 \leq j \leq k \leq n$ such that $s_j^{ap} \times e_k^{ap}$ is maximized. We then choose the start and end positions of the opinion span similarly such that they do not overlap with the aspect span. Thus, we obtain one set of four pointer probabilities. We repeat the process to obtain the second set, this time by choosing the opinion span before the aspect span. Finally, we choose the set (of aspect and opinion spans) that gives the higher product of the four probabilities.
\section{Experiments}\label{sec:experiments}
\subsection{Datasets and Evaluation Metrics}
We conduct our experiments on the \textbf{ASTE-Data-V2} dataset created by \newcite{jet-xu-etal-2020-position}. It is derived from \textbf{ASTE-Data-V1} \citep{Peng2020KnowingWH} and presents a more challenging scenario with \textbf{27.68\%} of all sentences containing triplets with overlapping aspect or opinion spans. The dataset contains triplet-annotated sentences from two domains: laptop and restaurant, corresponding to the original datasets released by the SemEval Challenge \citep{pontiki-etal-2014-semeval, pontiki-etal-2015-semeval, pontiki-etal-2016-semeval}. It is to be noted here that the opinion term annotations were originally derived from \citep{fan-etal-2019-target}. \textit{14Lap} belongs to the laptop domain and is henceforth referred to as the \textit{Laptop}. \textit{14Rest}, \textit{15Rest}, and \textit{16Rest} belong to the restaurant domain. Each dataset comes with its pre-defined split of training, development, and test sets. Similar to prior works, we report our results on the individual datasets. Additionally, we also conduct experiments on the combined restaurant dataset, henceforth referred to as the \textit{Restaurant}. Tables \ref{tab:ASTE-V2-trip} and \ref{tab:ASTE-V2-split} present the dataset statistics.

We consider \textit{precision}, \textit{recall}, and \textit{micro-F1} as our evaluation metrics for the triplet extraction task. A predicted triplet is considered a true positive only if all three predicted elements exactly match with those of a ground-truth opinion triplet.

\subsection{Experimental Setup}\label{subsec:setup}
For our non-BERT experiments, word embeddings are initialized (and kept trainable) using pre-trained 300-dim. Glove vectors \citep{pennington-etal-2014-glove}, and accordingly $d_w$ is set to 300. $d_{pos}$ and $d_{dep}$ are set to 50 each. $d_h$ is set to 300, and accordingly the hidden state dimensions of both the LSTMs (backward and forward) of the Bi-LSTM-based encoder are set to 150 each. $d_p$ is set to 300. For our BERT experiments, \emph{uncased} version of pre-trained BERT-base \citep{devlin-etal-2019-bert} is fine-tuned to encode each sentence. All our model variants are trained end-to-end on Tesla P100-PCIE 16GB GPU with \textit{Adam} optimizer (learning rate: $10^{-3}$, weight decay: $10^{-5}$). A \emph{dropout} rate of 0.5 is applied on the embeddings to avoid overfitting\footnote{Please refer to the appendix for more details.}. We make our codes and datasets publicly available\footnote{https://github.com/rajdeep345/PASTE/}.

\begin{table*}
    \footnotesize
    \centering
    \begin{tabular}{|l|c|c|c|c|c|c|c|c|}
        \hline
        \multirow{2}{*}{\textbf{Model}} &
        \multicolumn{4}{c|}{\textbf{Laptop}} &
        \multicolumn{4}{c|}{\textbf{Restaurant}} \\
        \cline{2-9}
        & \textbf{P.} & \textbf{R.} & \textbf{F\textsubscript{1}} & \textbf{Dev F\textsubscript{1}} & \textbf{P.} & \textbf{R.} & \textbf{F\textsubscript{1}} & \textbf{ Dev F\textsubscript{1}} \\
        \hline
        
        CMLA\textsuperscript{+} & 0.301 & 0.369 & 0.332 & - & - & - & - & -\\
        
        RINANTE\textsuperscript{+} & 0.217 & 0.187 & 0.201 & - & - & - & - & -\\
        
        Li-unified-R & 0.406 & 0.443 & 0.423 & - & - & - & - & - \\
        
        \citep{Peng2020KnowingWH} & 0.374 & 0.504 & 0.429 & - & - & - & - & - \\
        
        JET\textsuperscript{o} (M = 4) & 0.546 & 0.354 & 0.429 & 0.457 & 0.770 & 0.520 & 0.621 & 0.641 \\
        
        
        
        JET\textsuperscript{o} (M = 5) & 0.560 & 0.354 & 0.433 & 0.458 & - & - & - & -\\
        
        OTE-MTL & 0.492 & 0.405 & 0.451 & 0.458 & 0.710 & 0.579 & 0.637 & 0.729 \\
        
        
        GTS-BiLSTM w/o DE & 0.597 & 0.348 & 0.439 & 0.465 & 0.768 & 0.629 & 0.692 & 0.748 \\
        \hline
        
        PASTE-AF & 0.537 & 0.486 & \bf{0.510} & 0.496 & 0.707 & 0.701 & 0.704 & 0.741 \\
        
        PASTE-OF & 0.521 & 0.481 & 0.500 & 0.482 & 0.707 & 0.706 & \bf{0.707} & 0.740 \\
        \hline
        
        \multicolumn{9}{|l|}{\textbf{With BERT}} \\
        \hline
        
        JET\textsuperscript{o} (M = 4) & 0.570 & 0.389 & 0.462 & 0.475 & 0.727 & 0.549 & 0.626 & 0.645 \\
        
        JET\textsuperscript{o} (M = 6) & 0.554 & 0.473 & 0.510 & 0.488 & - & - & - & - \\
        
        GTS-BERT & 0.549 & 0.521 & 0.535 & 0.579 & 0.748 & 0.732 & 0.\bf{740} & 0.767 \\
        \hline
        
        PASTE-AF & 0.550 & 0.516 & 0.532 & 0.514 & 0.710 & 0.704 & 0.707 & 0.744 \\
        
        \quad w/ BERT-PT & 0.612 & 0.536 & 0.571 & 0.576 & 0.747 & 0.718 & 0.732 & 0.759 \\
        
        PASTE-OF & 0.554 & 0.519 & \bf{0.536} & 0.503 & 0.705 & 0.705 & 0.705 & 0.744 \\
        
        \quad w/ BERT-PT & 0.597 & 0.553 & \underline{\bf{0.574}} & 0.547 & 0.737 & 0.737 & \underline{0.737} & 0.759 \\
        \hline
        
    \end{tabular}
    \caption{Comparative results on the \emph{Laptop} (14Lap) and \emph{Restaurant} datasets from ASTE-Data-V2. \textbf{Bolded} values represent the best F\textsubscript{1} scores. \underline{Underlined} scores are obtained with Post-trained BERT.}
    \label{tab:main_lap_resall}
\end{table*}

\begin{table*}[!thb]
    \centering
    \resizebox{\linewidth}{!}{
    \begin{tabular}{|l|c|c|c|c|c|c|c|c|c|c|c|c|}
        \hline
        \multirow{2}{*}{\textbf{Model}} &
        \multicolumn{4}{c|}{\textbf{14Rest}} &
        \multicolumn{4}{c|}{\textbf{15Rest}} &
        \multicolumn{4}{c|}{\textbf{16Rest}} \\
        \cline{2-13}
        & \textbf{P.} & \textbf{R.} & \textbf{F\textsubscript{1}} & \textbf{Dev F\textsubscript{1}} & \textbf{P.} & \textbf{R.} & \textbf{F\textsubscript{1}} & \textbf{Dev F\textsubscript{1}} & \textbf{P.} & \textbf{R.} & \textbf{F\textsubscript{1}} & \textbf{Dev F\textsubscript{1}} \\
        \hline
        
        CMLA\textsuperscript{+} & 0.392 & 0.471 & 0.428 & - & 0.346 & 0.398 & 0.370 & - & 0.413 & 0.421 & 0.417 & - \\
        
        RINANTE\textsuperscript{+} & 0.314 & 0.394 & 0.350 & - & 0.299 & 0.301 & 0.300 & - & 0.257 & 0.223 & 0.239 & - \\
        
        Li-unified-R & 0.410 & 0.674 & 0.510 & - & 0.447 & 0.514 & 0.478 & - & 0.373 & 0.545 & 0.443 & - \\
        
        \citep{Peng2020KnowingWH} & 0.432 & 0.637 & 0.515 & - & 0.481 & 0.575 & 0.523 & - & 0.470 & 0.642 & 0.542 & - \\
        
        OTE-MTL & 0.630 & 0.551 & 0.587 & 0.547 & 0.579 & 0.427 & 0.489 & 0.569 & 0.603 & 0.534 & 0.565 & 0.597 \\
        
        
        JET\textsuperscript{o} $(M = 6)$ & 0.615 & 0.551 & 0.581 & 0.535 & 0.644 & 0.443 & 0.525 & 0.610 & 0.709 & 0.570 & \bf{0.632} & 0.609 \\
        
        
        GTS-BiLSTM w/o DE & 0.686 & 0.528 & 0.597 & 0.556 & 0.654 & 0.443 & 0.528 & 0.606 & 0.686 & 0.515 & 0.588 & 0.625 \\
        \hline
        
        PASTE-AF & 0.624 & 0.618 & 0.621 & 0.568 & 0.548 & 0.534 & \bf{0.541} & 0.649 & 0.622 & 0.628 & 0.625 & 0.667 \\
        
        PASTE-OF & 0.634 & 0.619 & \bf{0.626} & 0.566 & 0.548 & 0.526 & 0.537 & 0.650 & 0.623 & 0.636 & 0.629 & 0.659 \\
        \hline
        
        \multicolumn{13}{|l|}{\textbf{With BERT}} \\
        \hline
        
        JET\textsuperscript{o} (M = 6) & 0.706 & 0.559 & 0.624 & 0.569 & 0.645 & 0.520 & 0.575 & 0.648 & 0.704 & 0.584 & 0.638 & 0.638 \\
        
        GTS-BERT & 0.674 & 0.673 & \bf{0.674} & 0.651 & 0.637 & 0.551 & \bf{0.591} & 0.720 & 0.654 & 0.680 & \bf{0.667} & 0.715 \\
        \hline
        
        PASTE-AF & 0.648 & 0.638 & 0.643 & 0.570 & 0.583 & 0.567 & 0.575 & 0.626 & 0.655 & 0.644 & 0.650 & 0.660 \\
        
        \quad w/ BERT-PT & 0.667 & 0.665 & \underline{0.666} & 0.585 & 0.617 & 0.608 & 0.613 & 0.673 & 0.661 & 0.698 & \underline{0.679} & 0.690 \\
        
        PASTE-OF & 0.667 & 0.608 & 0.636 & 0.573 & 0.585 & 0.565 & 0.575 & 0.645 & 0.619 & 0.667 & 0.642 & 0.670 \\
        
        \quad w/ BERT-PT & 0.687 & 0.638 & 0.661 & 0.592 & 0.636 & 0.598 & \underline{0.616} & 0.660 & 0.680 & 0.677 & 0.678 & 0.695 \\
        \hline
        
    \end{tabular}
    }
    \caption{Comparative results on the individual restaurant datasets from ASTE-Data-V2}
    \label{tab:main_res}
\end{table*}

\subsection{Baselines}

\begin{itemize}[leftmargin=*]
    \item \newcite{cmla} (\textbf{CMLA}) and \newcite{rinante} (\textbf{RINANTE}) propose different methods to co-extract aspects and opinion terms from review sentences. \newcite{li2019unified} propose a unified tagging scheme-based method for extracting opinion target-sentiment pairs. \newcite{Peng2020KnowingWH} modifies these methods to jointly extract targets with sentiment, and opinion spans. It then applies an MLP-based classifier to determine the validity of all possible generated triplets. These modified versions are referred to as \textbf{CMLA\textsuperscript{+}}, \textbf{RINANTE\textsuperscript{+}}, and \textbf{Li-unified-R}, respectively.
    \item \newcite{Peng2020KnowingWH} propose a BiLSTM+GCN-based approach to co-extract aspect-sentiment pairs, and opinion spans. They then use the same inference strategy as above to confirm the correctness of the generated triplets.
    \item \textbf{OTE-MTL} \citep{otemtl-zhang-etal-2020-multi-task} uses a multi-task learning framework to jointly detect aspects, opinions, and sentiment dependencies.
    \item \textbf{JET} \citep{jet-xu-etal-2020-position} is the first end-to-end approach for the task of ASTE that leverages a novel position-aware tagging scheme. One of their variants, \textbf{JET\textsuperscript{t}}, however cannot handle aspect-overlapped triplets. Similarly, \textbf{JET\textsuperscript{o}}, cannot handle opinion-overlapped triplets.
    \item \textbf{GTS} \citep{wu-etal-2020-grid} models ASTE as a novel grid-tagging task. However, given that it predicts the sentiment relation between all possible word pairs, it uses a relaxed (majority-based) matching criteria to determine the final triplets.
\end{itemize}

\subsection{Experimental Results}
\vspace{-0.2em}
While training our model variants, the best weights are selected based on F\textsubscript{1} scores on the development set. We report our median scores over 5 runs of the experiment. Performance comparisons on the \textit{Laptop} (14Lap) and combined \textit{Restaurant} datasets are reported in Table \ref{tab:main_lap_resall}, whereas the same on individual restaurant datasets are reported in Table \ref{tab:main_res}. Both the tables are divided into two sections; the former comparing the results without BERT, and the latter comparing those with BERT. The scores for \textbf{CMLA\textsuperscript{+}}, \textbf{RINANTE\textsuperscript{+}}, \textbf{Li-unified-R}, and \citep{Peng2020KnowingWH} are taken from \newcite{jet-xu-etal-2020-position}. We replicate the results for \textbf{OTE-MTL} on ASTE-Data-V2 and report their average scores over 10 runs of the experiment. For \textbf{JET}, we compare with their best reported results on the individual datasets; i.e.  JET\textsuperscript{o} (M = 5) for 14Lap (w/o BERT), JET\textsuperscript{o} (M = 6) for 14Lap (w/ BERT), and JET\textsuperscript{o} (M = 6) for 14Rest, 15Rest, and 16Rest (both w/ and w/o BERT). However, owing to resource constraints and known optimization issues with their codes, we could not replicate their results on the \textit{Restaurant} dataset beyond M = 4 (for both w/ and w/o BERT). \textbf{GTS} uses double embeddings \citep{xu-etal-2018-double} (general Glove vectors + domain-specific embeddings trained with \emph{fastText}). For fair comparison, we replicate their results without using the domain-specific embeddings (DE). For both w/ and w/o BERT, we report their median scores over 5 runs of the experiment. We also report the F\textsubscript{1} scores on the development set corresponding to the test set results.
\begin{table*}
    \footnotesize
    \centering
    \begin{tabular}{|l|c|c|c|c|c|c|c|c|}
        \hline
        \multirow{2}{*}{\textbf{Model}} &
        \multicolumn{4}{c|}{\textbf{Laptop}} &
        \multicolumn{4}{c|}{\textbf{Restaurant}} \\
        \cline{2-9}
        & \textbf{Single} & \textbf{Multi} & \textbf{MultiPol} & \textbf{Overlap} & \textbf{Single} & \textbf{Multi} & \textbf{MultiPol} & \textbf{Overlap} \\
        \hline
        
        JET\textsuperscript{o} (M = 4) & 0.453 & 0.406 & 0.219 & 0.363 & 0.654 & 0.602 & 0.558 & 0.518 \\
        
        OTE-MTL & 0.485 & 0.277 & 0.172 & 0.380 & 0.716 & 0.656 & 0.506 & 0.646 \\
        
        GTS-BiLSTM w/o DE & 0.418 & 0.452 & \bf{0.237} & 0.403 & \bf{0.726} & 0.675 & \bf{0.588} & 0.660 \\
        \hline
        
        PASTE-AF & \bf{0.506} & \bf{0.512} & 0.216 & 0.507 & 0.702 & \bf{0.705} & 0.567 & 0.688 \\
        
        PASTE-OF & 0.495 & 0.502 & 0.205 & \bf{0.511} & 0.711 & 0.704 & 0.582 & \bf{0.693} \\
        \hline
        
        \multicolumn{9}{|l|}{\textbf{With BERT}} \\
        \hline
        
        JET\textsuperscript{o} (M = 4) & 0.514 & 0.430 & 0.229 & 0.400 & 0.655 & 0.609 & 0.509 & 0.536\\
        
        GTS-BERT & 0.533 & \bf{0.536} & \bf{0.338} & \bf{0.540} & \bf{0.739} & \bf{0.740} & \bf{0.648} & \bf{0.722} \\
        \hline
        
        PASTE-AF & 0.555 & 0.519 & 0.265 & 0.526 & 0.704 & 0.709 & 0.601 & 0.699 \\
        
        PASTE-OF & \bf{0.593} & 0.502 & 0.282 & 0.511 & 0.699 & 0.708 & 0.571 & 0.697 \\
        \hline 
        
    \end{tabular}
    \caption{Comparison of F1 scores on different splits of \emph{Laptop} and \emph{Restaurant} datasets from ASTE-Data-V2}
    \label{tab:analysis_lap_resall}
\end{table*}

From Table \ref{tab:main_lap_resall}, both our variants, PASTE-AF and PASTE-OF, perform comparably as we substantially outperform all the non-BERT baselines. On \emph{Laptop}, we achieve \textbf{13.1\%} F\textsubscript{1} gains over OTE-MTL, whereas on \emph{Restaurant}, we obtain \textbf{2.2\%} F\textsubscript{1} gains over GTS-BiLSTM. We draw similar conclusions from Table \ref{tab:main_res}, except that we are narrowly outperformed by JET\textsuperscript{o} (M = 6) on 16Rest. Our better performance may be attributed to our better \textit{Recall} scores with around \textbf{15.6\% recall gains} (averaged across both our variants) over the respective strongest baselines (in terms of F\textsubscript{1}) on the \textit{Laptop} and \textit{Restaurant} datasets. Such an observation establishes the better efficacy of PASTE in modeling the interactions between the three \emph{opinion factors} as we are able to identify more ground-truth triplets from the data, compared to our baselines.

With BERT, we comfortably outperform JET on all the datasets. Although we narrowly beat GTS-BERT on \emph{Laptop}, it outperforms us on all the restaurant datasets. This is owing to the fact that GTS-BERT obtains a substantial improvement in scores over GTS since its grid-tag prediction task and both the pre-training tasks of BERT are all discriminative in nature. We on the other hand, do not observe such huge jumps (F\textsubscript{1} gains of 5.1\%, 2.7\%, 6.3\%, and 3.3\% on the \emph{Laptop}, Rest14, Rest15, and Rest16 datasets respectively, noticeably more improvement on datasets with lesser training data; no gains on \emph{Restaurant}) since BERT is known to be unsuitable for generative tasks. We envisage to improve our model by replacing BERT with BART \citep{lewis-etal-2020-bart}, a strong sequence-to-sequence pretrained model for NLG tasks. 

Finally, motivated by \newcite{xu-etal-2019-bert, xu-etal-2020-dombert}, we also demonstrate the utility of leveraging domain-specific language understanding for the task by reporting our results with BERT-PT (task-agnostic post-training of pre-trained BERT on domain-specific data) in both the tables. While we achieve substantial performance improvement, we do not use these scores to draw our conclusions in order to ensure fair comparison with the baselines.


\begin{table*}[!thb]
    \footnotesize
    \centering
    \begin{tabular}{|l|l|c|c|c|c|c|c|c|}
        \hline
        \multirow{2}{*}{\textbf{Dataset}} & \multirow{2}{*}{\textbf{Model}} & \multicolumn{3}{c|}{\textbf{Aspect}} &
        \multicolumn{3}{c|}{\textbf{Opinion}} & \textbf{Sentiment} \\
        \cline{3-9}
        & & \textbf{P.} & \textbf{R.} & \textbf{F\textsubscript{1}} & \textbf{P.} & \textbf{R.} & \textbf{F\textsubscript{1}} & 
        \textbf{\% Acc.} \\
        \hline
        
        \multirow{5}{*}{\textbf{Laptop}} & JET\textsuperscript{o} (M = 4) & 0.801 & 0.495 & 0.611 & 0.805 & 0.528 & 0.638 & 0.846 \\ 
        
        & OTE-MTL & 0.812 & 0.576 & 0.674 & 0.826 & 0.584 & 0.684 & 0.858 \\
        
        & GTS-BiLSTM w/o DE & 0.725 & 0.724 & 0.724 & 0.692 & 0.684 & 0.688 & \bf{0.870} \\
        \cline{2-9}
        
        & PASTE-AF & 0.792 & 0.765 & 0.778 & 0.757 & 0.704 & 0.730 & 0.840 \\
        
        & PASTE-OF & 0.801 & 0.790 & \bf{0.796} & 0.763 & 0.719 & \bf{0.740} & 0.831  \\
        
        
        
        
        
        
        
        \hline
        
        \multirow{5}{*}{\textbf{Restaurant}} & JET\textsuperscript{o} (M = 4) & 0.871 & 0.638 & 0.736 & 0.885 & 0.666 & 0.760 & \bf{0.947} \\
        
        & OTE-MTL & 0.905 & 0.706 & 0.793 & 0.913 & 0.718 & 0.804 & 0.943 \\
        
        & GTS-BiLSTM w/o DE & 0.791 & 0.835 & 0.812 & 0.826 & 0.837 & 0.832 & 0.945 \\
        \cline{2-9}
        
        & PASTE-AF & 0.837 & 0.851 & \bf{0.844} & 0.844 & 0.852 & 0.848 & 0.939 \\
        
        & PASTE-OF & 0.836 & 0.848 & 0.842 & 0.848 & 0.854 & \bf{0.851} & 0.939 \\
        
        
        
        
        
        
        
        \hline
    \end{tabular}
    \caption{Comparative results of aspect, opinion and sentiment prediction on \emph{Laptop} and \emph{Restaurant} datasets}
    \label{tab:split_analysis}
\end{table*}

\begin{table}[!thb]
    \centering
    \resizebox{\columnwidth}{!}{
    \begin{tabular}{|l|l|c|c|c|c|}
        \hline
        \textbf{Dataset} & \textbf{Model} & \textbf{P.} & \textbf{R.} & \textbf{F\textsubscript{1}} & \textbf{\% F\textsubscript{1} $\downarrow$} \\
        \hline
        
        \multirow{3}{*}{\textbf{Laptop}} & PASTE-AF & 0.537 & 0.486 & 0.510 & - \\
        & $\;$ - POS \& DEP & 0.530 & 0.451 & 0.488 & 4.3\% \\
        & $\;$ w/ Random & 0.505 & 0.410 & 0.453 & 11.2\% \\
        
        \hline
        
        \multirow{3}{*}{\textbf{Restaurant}} & PASTE-OF & 0.707 & 0.706 & 0.707 & - \\
        & $\;$ - POS \& DEP & 0.708 & 0.702 & 0.705 & 0.3\% \\
        & $\;$ w/ Random & 0.686 & 0.627 & 0.655 & 7.4\% \\
        
        \hline
    \end{tabular}
    }
    \caption{Ablation Results}
    \label{tab:ablations}
\end{table}

\section{Analysis \& Discussion}\label{analysis}
\subsection{Robustness Analysis}\label{subsec:qual_analysis}
In order to better understand the relative advantage of our proposed approach when compared to our baselines for the opinion triplet extraction task, and to further investigate the reason behind our better recall scores, in Table \ref{tab:analysis_lap_resall} we compare the F\textsubscript{1} scores on various splits of the test sets as defined in Table \ref{tab:ASTE-V2-split}. We observe that with our core architecture (w/o BERT), PASTE consistently outperforms the baselines on both \textit{Laptop} and \textit{Restaurant} datasets when it comes to handling sentences with multiple triplets, especially those with overlapping aspect/opinion spans. This establishes the fact that PASTE is better than previous tagging-based approaches in terms of modeling aspect-opinion span-level interdependence during the extraction process. This is an important observation considering the industry-readiness \citep{myECIR} of our proposed approach since our model is robust towards challenging data instances. We however perform poorly when it comes to identifying triplets with varying sentiment polarities in the same sentence. This is understandable since we do not utilize any specialized sentiment modeling technique. In future, we propose to utilize word-level Valence, Arousal, Dominance scores \citep{mySIGIR2021} as additional features to better capture the sentiment of the opinion phrase.

In this work, we propose a new perspective to solve ASTE by investigating the utility of a tagging-free scheme, as against all prior tagging-based methods. Hence, it becomes imperative to analyze how we perform in terms of identifying individual elements of an opinion triplet. Table \ref{tab:split_analysis} presents such a comparison. It is encouraging to note that we substantially outperform our baselines on both aspect and opinion span detection sub-tasks. However, as highlighted before, we are outperformed when it comes to sentiment detection.

\subsection{Ablation Study: }\label{subsec:ablation}
Since our \textit{Decoder} learns to decode the sequence of triplets from left to right without repetition, while training our models we sort the target triplets in the same order as \textit{generation direction}; i.e. for training PASTE-AF/PASTE-OF, the target triplets are sorted in ascending order of aspect/opinion start positions. As an ablation, we sort the triplets randomly while training the models and report our obtained scores in Table \ref{tab:ablations}. An average drop of 9.3\% in F\textsubscript{1} scores for both our model variants establish the importance of sorting the triplets for training our models. When experimenting without the POS and DEP features, we further observe an average drop of 2.3\% in F\textsubscript{1} scores, thereby demonstrating their utility for the ASTE task. When experimenting with BERT, although these features helped on the \emph{Laptop} and Rest15 datasets, overall we did not observe any significant improvement.

\section{Related Works}\label{related}
ABSA is a collection of several fine-grained sentiment analysis tasks, such as \emph{Aspect Extraction} \citep{ijcai2018-583, li-etal-2020-conditional}, \emph{Aspect-level Sentiment Classification} \citep{li-etal-2018-transformation, xue-li-2018-aspect}, \emph{Aspect-oriented Opinion Extraction} \citep{fan-etal-2019-target}, \emph{E2E-ABSA} \citep{li2019unified, he-etal-2019-interactive}, and \emph{Aspect-Opinion Co-Extraction} \citep{cmla, rinante}. However, none of these works offer a complete picture of the aspects being discussed. Towards this end, \newcite{Peng2020KnowingWH} recently coined the task of Aspect Sentiment Triplet Extraction (ASTE), and proposed a 2-stage pipeline solution. More recent end-to-end approaches such as OTE-MTL\citep{otemtl-zhang-etal-2020-multi-task}, and GTS \citep{wu-etal-2020-grid} fail to guarantee sentiment consistency over multi-word aspect/opinion spans, since they depend on word-pair dependencies. JET \citep{jet-xu-etal-2020-position} on the other hand requires two different models to be trained to detect aspect-overlapped and opinion-overlapped triplets. Different from all these tagging-based methods, we propose a tagging-free solution for the ASTE task.
\section{Conclusion}\label{conclusion}
We investigate the utility of a tagging-free scheme for the task of Aspect Sentiment Triplet Extraction using a Pointer network-based decoding framework. Addressing the limitations of previous tagging-based methods, our proposed architecture, \textbf{PASTE}, not only exploits the aspect-opinion interdependence during the span detection process, but also models the span-level interactions for sentiment prediction, thereby truly capturing the inter-relatedness between all three elements of an opinion triplet. We demonstrate the better efficacy of PASTE, especially in \emph{recall}, and in predicting multiple and/or overlapping triplets, when experimenting on the \emph{ASTE-Data-V2} dataset.

\section*{Acknowledgements}
This research is supported by IMPRINT-2, Science and Engineering Research Board (SERB), India.

\newpage
\bibliography{main}
\bibliographystyle{acl_natbib}

\newpage
\appendix
\section{Appendix}\label{sec:appendix}
\subsection{Pointer Network-based Decoder}\label{subsec:decoder_supl}
Referring to Figure \ref{fig:model}, opinion triplets are decoded using an LSTM-based \textit{Triplet Decoder}, that takes into account the history of previously generated pairs/tuples of aspect and opinion spans, in order to avoid repetition. At each time step $t$, it generates a hidden representation $\boldsymbol{\mathrm{h_t^D}} \in \mathbb{R}^{d_h}$ that is used by the two Bi-LSTM + FFN-based \textit{Pointer Networks} to respectively predict the aspect and opinion spans, while exploiting their interdependence. The tuple representation $tup_t$ thus obtained is concatenated with $\boldsymbol{\mathrm{h_t^D}}$ and passed through an FFN-based \textit{Sentiment Classifier} to predict the connecting sentiment, thereby decoding an entire opinion triplet at the $t^{th}$ time step. We now elaborate each component of our proposed decoder framework in greater depth.

\subsubsection{Span Detection with Pointer Networks}\label{subsubsec:span-detecion}
Our pointer network consists of a Bi-LSTM, with hidden dimension $d_p$, followed by two feed-forward layers (FFN) on top to respectively predict the start and end locations of an entity span. We use two such pointer networks to  produce a tuple of hidden vectors corresponding to the aspect and opinion spans of the triplet to be decoded at time step $t$. We concatenate $\boldsymbol{\mathrm{h_t^D}}$ with each of the encoder hidden state vectors $\boldsymbol{\mathrm{h_i^E}}$ and pass them as input to the first Bi-LSTM. The output hidden state vector corresponding to the $i^{th}$ token of the sentence thus obtained is simultaneously fed to the two FFNs with \textit{sigmoid} to generate a pair of scores $\tilde{s}_i^{p_1}$ and $\tilde{e}_i^{p_1}$ in the range of 0 to 1 as follows: 
\begin{equation}
    \tilde{s}_i^{p_1} = \boldsymbol{\mathrm{W_s^{p_1} h_i^{p_1}}} + \boldsymbol{\mathrm{b_s^{p_1}}}, \quad \tilde{e}_i^{p_1} = \boldsymbol{\mathrm{W_e^{p_1} h_i^{p_1}}} + \boldsymbol{\mathrm{b_e^{p_1}}} \nonumber
\end{equation}
Here, $\boldsymbol{\mathrm{W_s^{p_1}}} \in \mathbb{R}^{d_p \times 1}$, $\boldsymbol{\mathrm{W_e^{p_1}}} \in \mathbb{R}^{d_p \times 1}$, $\boldsymbol{\mathrm{b_s^{p_1}}}$, and $\boldsymbol{\mathrm{b_e^{p_1}}}$ are respectively the weights and bias parameters of the two FFNs for the first pointer network ($p_1$). After repeating the process for all tokens in the sentence, the normalized probabilities of the $i^{th}$ token to be the start and end positions of an \textit{aspect span} ($s_i^{p_1}$ and $e_i^{p_1}$ respectively) are obtained using \textit{softmax} operations over the two sets of scores thus generated (by the two FFNs) as follows: 
\begin{equation}
    \mathrm{S^{p_1}} = \mathrm{softmax(\tilde{S}^{p_1})}, \quad \mathrm{E^{p_1}} = \mathrm{softmax(\tilde{E}^{p_1})} \nonumber
\end{equation}
Similar equations are used for the second pointer network ($p_2$) to generate the normalized probabilities, $s_i^{p_2}$ and $e_i^{p_2}$, for the $i^{th}$ token to be the start and end positions of an opinion span respectively; difference being that apart from concatenating $\boldsymbol{\mathrm{h_t^D}}$, we also concatenate the output vectors $\boldsymbol{\mathrm{h_i^{p_1}}}$ from the first Bi-LSTM with encoder hidden states $\boldsymbol{\mathrm{h_i^E}}$ and pass them as input to the second Bi-LSTM. The vector representations for the aspect and opinion spans at time step $t$ are obtained as follows:
\begin{align*}
    \mathrm{ap_t} = \sum_{i=1}^n s_i^{p_1} \boldsymbol{\mathrm{h_i^{p_1}}} \: \mathbin\Vert \: \sum_{i=1}^n e_i^{p_1} \boldsymbol{\mathrm{h_i^{p_1}}}; \mathrm{ap_t} \in \mathbb{R}^{2d_p} \\
    \mathrm{op_t} = \sum_{i=1}^n s_i^{p_2} \boldsymbol{\mathrm{h_i^{p_2}}} \: \mathbin\Vert \: \sum_{i=1}^n e_i^{p_2} \boldsymbol{\mathrm{h_i^{p_2}}}; \mathrm{op_t} \in \mathbb{R}^{2d_p} \nonumber
\end{align*}

Here we introduce the term \textit{generation direction} which refers to the order in which we generate the hidden representations for the two entities, i.e. aspect and opinion spans. This allows us to define two \textbf{variants} of our model. The variant discussed so far uses $p_1$ to detect the aspect span before predicting the opinion span using $p_2$, and is henceforth referred to as \textbf{PASTE-AF} (AF stands for \textit{aspect first}). Similarly, we obtain the second variant \textbf{PASTE-OF} (\textit{opinion first}) by reversing the \textit{generation direction}. The other two components of our model remain the same for both the variants.

\subsection{Attention Modeling}\label{subsec:attention-supl}
\noindent We use \textit{Badhanau Attention} \citep{BadhanauAttention} to obtain the context representation of the input sentence ($s_t^E \in \mathbb{R}^{d_h}$) at time step $t$ as follows:
\begin{gather*}
    \tilde{tup}_{prev} = \boldsymbol{\mathrm{W_{tup}}} \; tup_{prev} + \boldsymbol{\mathrm{b_{tup}}} \\
    \boldsymbol{\mathrm{u_t^i}} = \boldsymbol{\mathrm{W_u}} \boldsymbol{\mathrm{h_i^E}} \\
    \boldsymbol{\mathrm{\tilde{q}_t^i}} = \boldsymbol{\mathrm{W_{\tilde{q}}}} \; \tilde{tup}_{prev} + \boldsymbol{\mathrm{b_{\tilde{q}}}} \:;\: \boldsymbol{\mathrm{\tilde{a}_t^i}} = \boldsymbol{\mathrm{v_{\tilde{a}}}} \: \mathrm{tanh} (\boldsymbol{\mathrm{\tilde{q}_t^i}} + \boldsymbol{\mathrm{u_t^i}}) \\
    \boldsymbol{\mathrm{q_t^i}} = \boldsymbol{\mathrm{W_q}} \boldsymbol{\mathrm{h_{t-1}^D}} + \boldsymbol{\mathrm{b_q}} \:;\: \boldsymbol{\mathrm{a_t^i}} = \boldsymbol{\mathrm{v_a}} \: \mathrm{tanh} (\boldsymbol{\mathrm{q_t^i}} + \boldsymbol{\mathrm{u_t^i}}) \\
    \boldsymbol{\tilde{\alpha}_t} = \mathrm{softmax} (\boldsymbol{\mathrm{\tilde{a}_t}}) \:;\: \boldsymbol{\alpha_t} = \mathrm{softmax} (\boldsymbol{\mathrm{a_t}}) \\
    \boldsymbol{\mathrm{s_t^E}} = \sum_{i=1}^n \frac{\tilde{\alpha}_i + \alpha_i}{2} \: \boldsymbol{\mathrm{h_i^E}}
\end{gather*}
Here, $\boldsymbol{\mathrm{W_{\tilde{q}}}}, \boldsymbol{\mathrm{W_{q}}}, \boldsymbol{\mathrm{W_{u}}} \in \mathrm{R}^{d_h \times d_h}$, $\boldsymbol{\mathrm{v_{\tilde{a}}}}, \boldsymbol{\mathrm{v_a}} \in \mathrm{R}^{d_h}$ are learnable attention parameters, and $\boldsymbol{\mathrm{b_{\tilde{q}}}}, \boldsymbol{\mathrm{b_{q}}} \in \mathrm{R}^{d_h}$ are bias vectors. First, we obtain $\tilde{tup}_{prev}$ from $tup_{prev}$ using a linear embedding layer, with $\boldsymbol{\mathrm{W_{tup}}} \in \mathrm{R}^{4d_p \times d_h}$ and $\boldsymbol{\mathrm{b_{tup}}}$ as its weights and bias parameters. We then use both $\tilde{tup}_{prev}$ and $\boldsymbol{\mathrm{h_{t-1}^D}}$ separately to obtain two attentive context vectors, $\boldsymbol{\mathrm{\tilde{q}_t}}$ and $\boldsymbol{\mathrm{q_t}}$ respectively. These are then concatenated along with $tup_{prev}$ to define the current context of our LSTM-based decoder. The corresponding normalized attention scores, $\boldsymbol{\tilde{\alpha}_t}$ and $\boldsymbol{\alpha_t}$, are averaged to obtain the attention-weighted sentence representation at decoding time step $t$. 

\subsection{Experimental Setup}\label{subsec:setup-supl}
For our non-BERT experiments, word embeddings are initialized (and kept trainable) using pre-trained 300-dim. Glove vectors \citep{pennington-etal-2014-glove}, and accordingly $d_w$ is set to 300. The dimensions of POS and DEP embeddings, i.e. $d_{pos}$ and $d_{dep}$ are set to 50 each. The decoder (LSTM) hidden dimension $d_h$ is set to 300, and accordingly the hidden state dimensions of both backward and forward LSTMs of the Bi-LSTM-based encoder are set to 150 each. We set the hidden dimension $d_p$ of the Bi-LSTMs in pointer networks to 300. For our BERT experiments, \emph{uncased} version of pre-trained BERT-base \citep{devlin-etal-2019-bert} is fine-tuned to encode each sentence.

All our model variants are trained end-to-end with \textit{Adam} optimizer \citep{Kingma2015AdamAM} with $10^{-3}$ as the learning rate, and $10^{-5}$ as weight decay. Dropout (0.5) \citep{dropout} is applied on embeddings to avoid overfitting. Our non-BERT model variants are trained for 100 epochs with a batch size of 10. Our BERT-based variants are trained for 30 epochs with a batch size of 16. Model selected according to the best F\textsubscript{1} score on the development data is used to evaluate on the test data. We run each model five times and report the median scores. All our experiments are run on Tesla P100-PCIE 16GB GPU.


\end{document}